%% file: arxiv_version.tex
\pdfoutput=1

\documentclass[11pt]{article}

\usepackage[final]{acl}

\usepackage{times}
\usepackage{latexsym}

\usepackage[T1]{fontenc}

\usepackage[utf8]{inputenc}

\usepackage{microtype}

\usepackage{inconsolata}

\usepackage{graphicx}
\usepackage{todonotes}
\usepackage{subfiles}
\usepackage{makecell}
\usepackage{subcaption}
\usepackage{spverbatim}
\usepackage{multirow}
\usepackage{rotating}
\usepackage{makecell}  
\usepackage{float} 

\usepackage{listings}
\usepackage{color}

\title{Ad-hoc Concept Forming in the Game Codenames\\ as a Means for Evaluating Large Language Models}

\author{
 \textbf{Sherzod Hakimov\textsuperscript{1}},
 \textbf{Lara Pfennigschmidt\textsuperscript{1}},
 \textbf{David Schlangen\textsuperscript{1,2}}
\\
\\
 \textsuperscript{1}Computational Linguistics, Department of Linguistics\\University of Potsdam, Germany\\
 \textsuperscript{2}German Research Center for Artificial Intelligence (DFKI), Berlin, Germany
\\
\texttt{firstname.lastname@uni-potsdam.de} 
}

\begin{document}
\maketitle
\begin{abstract}

This study utilizes the game Codenames as a benchmarking tool to evaluate large language models (LLMs) with respect to specific linguistic and cognitive skills. LLMs play each side of the game, where one side generates a clue word covering several target words and the other guesses those target words. We designed various experiments by controlling the choice of words (abstract vs.\ concrete words, ambiguous vs.\ monosemic) or the opponent (programmed to be faster or slower in revealing words). Recent commercial and open-weight models were compared side-by-side to find out factors affecting their performance. The evaluation reveals details about their strategies, challenging cases, and limitations of LLMs.

\end{abstract}

\section{Introduction}

The astounding abilities of large language models (LLMs) have led to what could be called a `crisis of evaluation', where the previous paradigm of evaluating natural language processing (NLP) models---through pairs of problem instance and expected response---does not fit well any more. First, the main mode of usage of LLMs is through their embedding in a ``chatbot'', often across multiple turns, which is not represented by the reference-based evaluation mode. Second, the closed nature and sheer size of the training data, often acquired through automatic means from the open internet, raises fears that the usual test datasets have been ingested and hence the training data has become \textit{contaminated}, rendering the value of the tests even more doubtful \cite{DBLP:conf/iclr/GolchinS24, deng-etal-2024-investigating}.

\begin{figure}[ht!]
    \centering
    \includegraphics[width=1.0\linewidth]{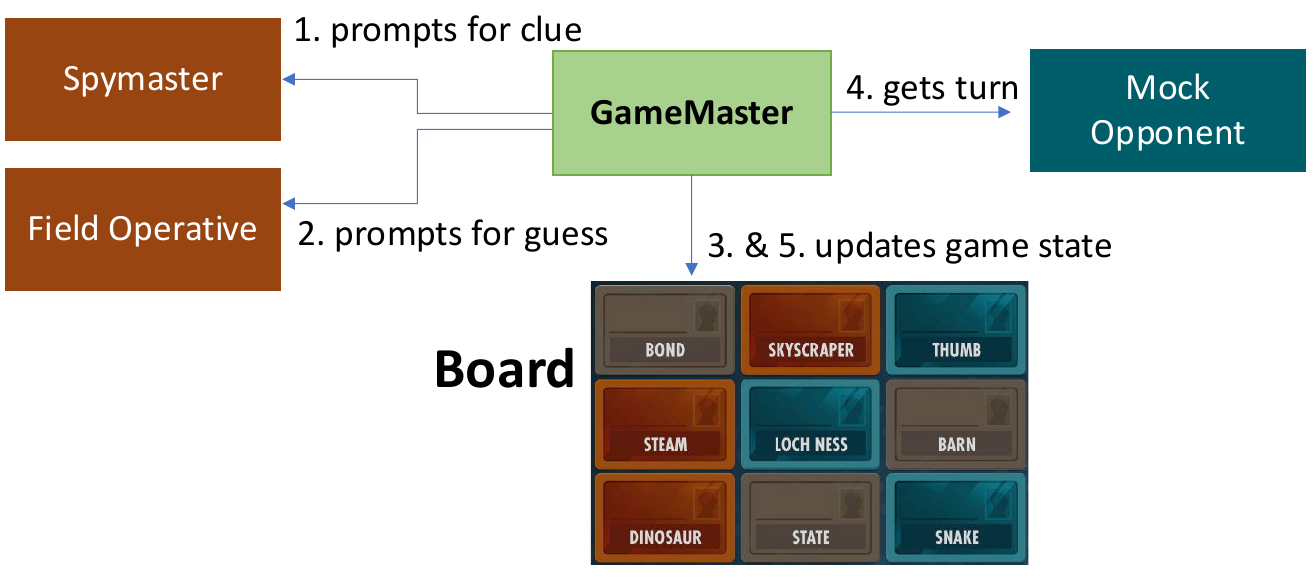}
    \caption{Overview of the proposed approach where an LLM poses as both Spymaster (clue giver) and Field Operative (guesser) and plays against a mock opponent. The GameMaster orchestrates the game play by keeping and updating the game state on the board (image from \textit{https://codenames.game/}.}
    \label{fig:overview}
\end{figure}

The use of games as an interactive environment where LLMs are tasked to perform certain actions and scored whether they achieve the task or not has emerged as a response to this situation~\cite{chalamalasetti-etal-2023-clembench, DBLP:conf/iclr/WuTML24, DBLP:conf/iclr/Zhou0MZYQMBFNS24}, allowing the evaluation to more closely approximate the interactive use situation, and overcoming the data contamination problem in two ways. First, even for already documented games, it is easy to create new instances that lead to game play that differs from what is in existing data \cite{beyer2024clembench2024challengingdynamiccomplementary}. Second, to further extend the range of evaluated phenomena, it is only necessary to add new game implementations, rather than to create new datasets. It is this second dimension that we explore in this paper.

We have implemented the game Codenames~\footnote{Source code ("codenames" directory): \url{https://github.com/clembench/clembench}} as a challenging means (already investigated in psycholinguistic studies, see below) for evaluating certain language-use capabilities. In this game, a first player needs to provide to a second player a clue which singles out certain words within a larger set of words given to both players (see Figure~\ref{fig:qual2} below for examples).

The game requires cooperation, with the first player needing to form and name an \textit{ad-hoc concept} that spans the target concepts, in a way that they assume is understandable to others (\textit{theory of mind}) \citep{kim-2019-understanding-NLP-via-codenames}.
Players need to connect words in a wide \textit{variety of relations} such as homonyms, antonyms, rhymes, or popular culture references \citep{jaramillo-2020-transformers-for-word-association-in-codenames}. Clue generation is also  a task of \textit{co-creativity} \citep{spendlove-constraints-for-creativity}, testing skills in the evaluation of \textit{semantic relatedness of word}s and \textit{common-sense reasoning} \citep{bitton-2022-winogavil} as well as the ability to constrain clues and negatively associate them with any non-team word.

Players need to predict the partners' behaviour and knowledge \citep{cserhati-2022-codenames-co-occurrence-counting, kumar-2021-connector-asociative-vs-distributional-semantic-models}, so one cannot simply optimise their own behaviour \citep{jaramillo-2020-transformers-for-word-association-in-codenames} without acknowledging the cultural background and knowledge level of their teammates \citep{shaikh-2023-cross-cultural-pragmatic-inference-with-codenames-duet}, hence requiring cooperation. Figure~\ref{fig:overview} shows the overall idea where players (Spymaster and Field Operative) are LLMs that play against a programmed mock opponent. The programmatic \textit{GameMaster} component (part of the framework we use) orchestrates the game play by providing inputs and generated outputs among the parties, checks whether players comply to the rules of the game.

Our contributions are as follows: i) benchmarking LLMs to test their ad-hoc concept generation, cooperation, pragmatic reasoning capabilities, ii) comparison of open-weight and commercial models under various experiments, iii) in-depth analysis of how best-performing models navigate the task.

\section{Related Work}

Earlier work focused on using various word embedding techniques (choose the clue that is closest to targets and most distant to distractors)~\cite{kim-2019-understanding-NLP-via-codenames,jaramillo-2020-transformers-for-word-association-in-codenames}. Later, such methods were combined with associative methods that use language graphs for generating clues or guessing~\cite{koyyalagunta-2021-codenames-language-graphs}. Other approaches involve concentrating on word co-occurrence measurements ~\cite{deRijk-2020-paper-word-embeddings-collocations, cserhati-2022-codenames-co-occurrence-counting} for capturing synonymy, semantic similarity, or word-relatedness measure instead of just focusing on word embeddings.

Later research started looking into using LLMs to generate clues~\cite{spendlove-2022-competitive-language-games-as-creative-tasks}. The idea of benchmarking LLMs led to the development of various datasets, e.g.\ BigBench~\cite{srivastava-2022-bigbench} includes Codenames as one of the many tasks to test emergent abilities of models~\cite{wei-2022-emergence,ozturkler-2023-thinksum,lu-2023-emergence-just-in-context-learning}.  \citet{DBLP:journals/corr/abs-2412-11373} recently explored using Codenames to benchmark LLMs where two pairs of LLMs (red vs.\ blue team) play against each other. Our method differs from theirs by comparing the language model team against a deterministic opponent. Not using a deterministic opponent could lead to different results every time the same game is played due to the non-deterministic nature of language models~\cite{song2024goodbadgreedyevaluation}. Another extension of our work lies in the experimental setup, where we study the effect of selecting words on a board and their relations in much more depth.

\section{The game: Codenames}\label{sec:codenames}

Codenames~\citep{chvatil-2015-codenames-game} is a cooperative board game with two teams (blue and red) that try to uncover their team agents' code names before the other team finds all of theirs. The board is set up with 25~word cards. Each team has a ``Spymaster'' that knows which words on the board represent their team (8+1 for the starting team), the opposing team (8), innocent bystanders (7), and the assassin (1). The team starting the game has one more word to uncover to balance out the advantage of going first. Our implemented version deploys one Spymaster and Field Operative on the same team. The opponent team is \textit{mocked} with an ideal behaviour of revealing \textit{n} own words each turn.

The Spymaster takes turns providing clues for their teammates -- the ``Field Operatives''. Each clue consists of a word related to one or more (code names) targeted words. It has to output in this way:
\begin{verbatim}
    CLUE: <clue>
    TARGETS: <list of targets>
\end{verbatim}

Only the clue is passed on to the Field Operative, who then guesses the matching words:
\begin{verbatim}
    GUESS: <list of guesses>
\end{verbatim}

If the guess is correct, the team can continue guessing as many names as the Spymaster indicated in their clue. If the team is unsure, they can also end their turn voluntarily. If the team's guess is incorrect, meaning they contacted an innocent bystander or an word of the opposing team, the identity is revealed, and the team's turn ends. If the team uncovers any \textit{assassin} word, the team immediately loses the game.

We have implemented the game using the clembench~\citep{chalamalasetti-etal-2023-clembench} framework where the GameMaster orchestrates the gameplay by 1) checking the required formatting of generated outputs by Spymaster or Field Operative, 2) passing the outputs between players. The Spymaster and Field Operative prompts are given in Appendix~\ref{appendix:prompts}.

\section{Experimental Setup}

\subsection{Board Generation}\label{subsec:board_generation}

We used different sets of words to design experiments. Each experiment includes 10 instances (boards) where the words are chosen randomly from a specific set. The default mock opponent uncovers one word per turn (\textit{n=1}). The default word list is by \citet{deRijk-2020-thesis-codenames-modelling-word-association} with \textit{one assassin word} per board. We defined the following experiments by changing specific default parameters, which correspond to 130 instances:



\noindent $\bullet$
\textbf{Risk level}: We included five assassin words in the set called \textit{high risk}. The \textit{low risk} set has no assassin words. The rationale here is to see whether models target less number of words to mitigate the risk of revealing assassin words.
    
\noindent $\bullet$
 \textbf{Word association}: We selected 45 category norms (e.g.\ bird name, kitchen utensil, country, military title, etc.) from the corpus by \citet{castro2021category-norms}. The \textit{easy} set is created by selecting 3-5 categories, sampling words for each category, and assigning them to the same team (3-5 turns by targeting the category). The \textit{difficult} set is created by ensuring that sampled words are distributed across all possible groups (team, opponent, innocent, assassin) and not assigned to the same team. The rationale here is whether models actually can capture those obvious associations on the easy set and whether they can play the difficult one at all.
    
\noindent $\bullet$
 \textbf{Opponent level}: We created three sets where the mock opponent turns \textit{two}, \textit{one} or \textit{none} words per turn, which correspond to \textit{difficult}, \textit{easy}, and \textit{none} levels, respectively. The rationale here is to check whether LLMs can play against a faster opponent that constantly reveals two words at a time.

\noindent $\bullet$
 \textbf{Word frequency}: All nouns from the SUBTLEX-US corpus~\cite{subtlexus} were filtered out to create two sets for low and high-frequency lists. We used the top and bottom 250 words for the frequency lists of the \textit{high} and \textit{low}. Typical human players would usually struggle with low frequency words and our rationale is to check whether it poses a similar challenge to LLMs too.
    
\noindent $\bullet$
 \textbf{Word ambiguity}: The corpus provided by \citet{beekhuizen2021ambiguity} includes monosemes (words with single sense) and homonyms (words with multiple senses). The \textit{ambiguous} set is composed of homonyms while the \textit{unambiguous} one includes the monosemes. The hypothesis here is that words with multiple meanings are easier to find connections between them than ambiguous words.
    
\noindent $\bullet$
 \textbf{Word concreteness}: Two sets of words where one corresponds to concrete concepts and the other includes abstract ones. \citet{brysbaert2014concreteness} collected word concreteness ratings (Likert scale between 1-5). We used the top 500 words with the lowest and highest concreteness ratings for \textit{abstract} and \textit{concrete} word lists, respectively. The hypothesis here to check whether LLMs play better with concrete words as it is easier for human players to find association between them in contrast to abstract concepts.

\subsection{Metrics}\label{subsec:metrics}

The clembench framewor measures how many of the instances (boards) have resulted in a \textit{Played} or \textit{Aborted} state. The gameplay is marked as \textit{Aborted} if either player does not follow the formatting instructions when generating an output (as explained in Section~\ref{sec:codenames}). \textit{Played} is the ratio of remaining gameplays (\textit{episodes}) where formatting instructions have been followed. The \textit{Played} ratio is further divided into \textit{Success} if the team reveals own words faster than mock opponent, or \textit{Lose}    if an assassin word is revealed or the mock is faster. 

The framework also requires one metric called \textit{Quality Score} corresponding to how well the task has been solved. The \textit{Quality Score}, essentially a win rate,  is the average number of games won (successful). The main ranking score for evaluated LLMs is the \textit{clemscore}, which is the macro-average quality score multiplied by the macro-average proportion of played games to find a balance between solving most tasks and following instructions. We have also implemented the following metrics to analyse the strategies taken by models: 

\noindent $\bullet$
\textbf{Sensitivity}: The number of revealed divided by the total team words.

\noindent $\bullet$
\textbf{Efficiency}: We set the bar at two target words per turn as the highest efficiency a model can reach, as that is a reasonable efficiency for humans. It is calculated as:\\
    $ min(1, \frac{1}{2}\cdot\frac{\mbox{team words revealed}}{\mbox{number of turns}})$

\subsection{Evaluated Models}
We evaluated open-weight and commercial models with a \textit{zero-shot} setting where \textit{temp=0}. We included the most recent commercial models such as: \textit{o3-mini} (Jan~'25), \textit{GPT-4o} (Aug~'24) \textit{Claude-3-5} (Sonnet, Oct~'24), and \textit{Gemini-2.0-Flash} (Feb~'25). We also included recent open-weight models: \textit{Llama-3.1} (8B, 70B, 405B)~\citep{llama31}, \textit{Llama-3.3} (70B), \textit{Qwen2} (72B)~\citep{qwen2}, \textit{Qwen2.5} (Coder-32B, 72B, Max)~\citep{qwen25}, and \textit{Deepseek} (v3, r1)~\citep{deepseekv3,deepseekai2025deepseekr1incentivizingreasoningcapability}. We used the APIs of the respective commercial models. For open-weight models, we ran the inference on two NVIDIA A100 GPUs. Two Deepseek models, Llama-3.1-405B and Qwen-Max, were run via the OpenRouter API.

\section{Results}

\begin{table}[]
\centering
\footnotesize
\begin{tabular}{l|c|c|c}
\textbf{Model} & \textbf{\makecell{clemscore}} & \textbf{\% Played} & \textbf{\makecell{Quality \\ Score}} \\ \hline

o3-mini & \textbf{49.2} & \textbf{100.0} & 49.2 \\
Claude-3-5 & 46.9 & 93.8 & 50.0 \\
GPT-4o & 45.4 & 93.8 & 48.4 \\
Deepseek-r1 & 45.4 & 85.4 & \textbf{53.2} \\
Gemini-2.0 & 37.7 & 96.2 & 39.2 \\
Llama-3.1-70B & 36.9 & 90.0 & 41.0 \\
Deepseek-v3 & 33.8 & 86.9 & 38.9 \\
Qwen2.5-72B & 30.0 & 72.3 & 41.5 \\
Llama-3.3-70B & 29.2 & 80.0 & 36.5 \\
Llama-3.1-405B & 29.2 & 76.2 & 38.4 \\
Qwen-max & 25.4 & 70.0 & 36.3 \\
Qwen2-72B & 20.8 & 58.5 & 35.5 \\
Qwen2.5-32B & 20.8 & 62.3 & 33.3 \\
Llama-3.1-8B & 14.6 & 52.3 & 27.9 \\
 \hline

\end{tabular}
\caption{Ranking of all benchmarked LLMs.}
\label{tab:main-results}
\end{table}

\begin{table*}[ht!]
\centering
\footnotesize

\begin{tabular}{|ll|c|c|c|c|c|c|c|}
\hline
\multicolumn{2}{|c|}{\textbf{Experiment}} & \textbf{o3-mini} & \textbf{GPT-4o} & \textbf{LM-3.1} & \textbf{LM-3.3} & \textbf{Claude-3.5} & \textbf{Deepseek-r1} & \textbf{Gemini-2.0} \\ \hline
\multicolumn{1}{|l|}{\multirow{2}{*}{Risk Level}}
& low & 70.0 & 75.0 & 50.0 & 50.0 & 75.0 & \textbf{87.5} & 55.6 \\ 
\multicolumn{1}{|l|}{}
& high & 20.0 & \textbf{37.5} & 11.1 & 20.0 & 30.0 & 10.0 & 10.0 \\ \hline
\multicolumn{1}{|l|}{\multirow{2}{*}{Association}}
& easy & \textbf{100.0} & \textbf{100.0} & \textbf{100.0} & \textbf{100.0} & \textbf{100.0} & \textbf{100.0} & \textbf{100.0} \\ 
\multicolumn{1}{|l|}{}
& difficult & 20.0 & 10.0 & 20.0 & 0.0 & 20.0 & \textbf{28.6} & 12.5 \\ \hline
\multicolumn{1}{|l|}{\multirow{3}{*}{Opponent}}
& none & \textbf{80.0} & 77.8 & \textbf{80.0} & 75.0 & 57.1 & 62.5 & 77.8 \\ 
\multicolumn{1}{|l|}{}
& easy & 50.0 & 33.3 & 14.3 & 28.6 & 40.0 & \textbf{80.0} & 11.1 \\ 
\multicolumn{1}{|l|}{}
& difficult & 0.0 & 0.0 & 0.0 & 0.0 & 0.0 & \textbf{22.2} & 0.0 \\ \hline
\multicolumn{1}{|l|}{\multirow{2}{*}{Frequency}}
& low & 60.0 & \textbf{66.7} & 50.0 & 33.3 & 60.0 & 50.0 & 30.0 \\ 
\multicolumn{1}{|l|}{}
& high & 20.0 & 30.0 & 50.0 & 20.0 & 44.4 & 25.0 & \textbf{50.0} \\ \hline
\multicolumn{1}{|l|}{\multirow{2}{*}{Ambiguity}}
& none & \textbf{80.0} & 60.0 & 22.2 & 55.6 & \textbf{80.0} & 55.6 & 40.0 \\ 
\multicolumn{1}{|l|}{}
& ambiguous & 40.0 & 33.3 & 37.5 & 10.0 & \textbf{62.5} & 16.7 & 40.0 \\ \hline
\multicolumn{1}{|l|}{\multirow{2}{*}{Concreteness}}
& concrete & 80.0 & 50.0 & 66.7 & 44.4 & 50.0 & \textbf{88.9} & 40.0 \\ 

\multicolumn{1}{|l|}{}
& abstract & 20.0 & \textbf{60.0} & 0.0 & 16.7 & 40.0 & 50.0 & 40.0 \\  \hline
\end{tabular}

\caption{Detailed results across different experiments. Only high performing LLMs were selected. The values correspond to the Quality Score for each experiment. \textit{LM-3.1} $\rightarrow$ Llama-3.1-405B, \textit{LM-3.3} $\rightarrow$ Llama-3.3-70B}
\label{tab:experiment-results}
\end{table*}

\subsection{Overall Analysis}
The overall results are given in Table~\ref{tab:main-results} where the \textit{clemscore}, \textit{Played}, and \textit{Quality Score} are averaged across all experiments. The first observation we make is that, as expected,larger models perform better. In line with this, commercial models outperform open-weight ones by some margin (five points between \textit{o3-mini} and \textit{Deepseek-r1}). \textit{o3-mini} is the only model that played all episodes without once making an instruction following error in the game. However, we can see that the best model achieves only \textit{49.2\%} success rate in winning the game against the mock opponent. To investigate specific experiments, we selected seven high-performing models to compare them in detail. The results are given in Table~\ref{tab:experiment-results}.

\textbf{Risk level}: We expected the high risk to be more complex than the low one because there are five assassin words. This expected behaviour holds for all models, e.g. \textit{o3-mini} has a margin of 50 points between both experiments. In the high-risk experiment, \textit{GPT-4o} achieves the best score of \textit{37.5}, which is a substantial margin of \textit{17.5} points compared to the second-best result.

\textbf{Word association}: All models achieved a perfect score for the \textit{easy} set. The difficult case is much more challenging as no model reaches 30 points.

\textbf{Opponent level}: We tested three levels of the mock opponent where the difference lies in how fast the words are revealed. The performance on the first level is significantly higher for all models as it is easier to beat the mock opponent who does not reveal any words. Even in this setting, the best models (\textit{o3-mini} and \textit{Llama-3.1-405B}) can only reach \textit{80} points. However, once we switch to other levels, we see a clear drop in performance for most models, except \textit{Deepseek-r1}. The difficult level shows even striking results where only \textit{Deepseek-r1} managed to achieve some performance while other models lost all episodes to the mock opponent.

\textbf{Word frequency}: The expectation here is that higher-frequency words are easier to play with (at least for human players). This assumption does not apply as most models are better at \textit{low frequency} set, except \textit{Gemini-2.0}.

\textbf{Ambiguity}: The expectation here is that monosemic words are easier to play with, and we can confirm that this holds for most models. \textit{Claude-3.5} is the only model to surpass 50\% success rate in the \textit{ambiguous} set.

\textbf{Concreteness}: Generally, all models perform better on the \textit{concrete} set, except for \textit{GPT-4o}. Interestingly, \textit{Gemini-2.0} gets equal points on both sets. It indicates that abstract words are indeed more challenging (as for humans) for models.

\subsection{In-depth Analysis}


\begin{figure}[ht!]
    \centering
    \begin{minipage}{0.45\textwidth}
        \centering
        \includegraphics[width=1.0\linewidth]{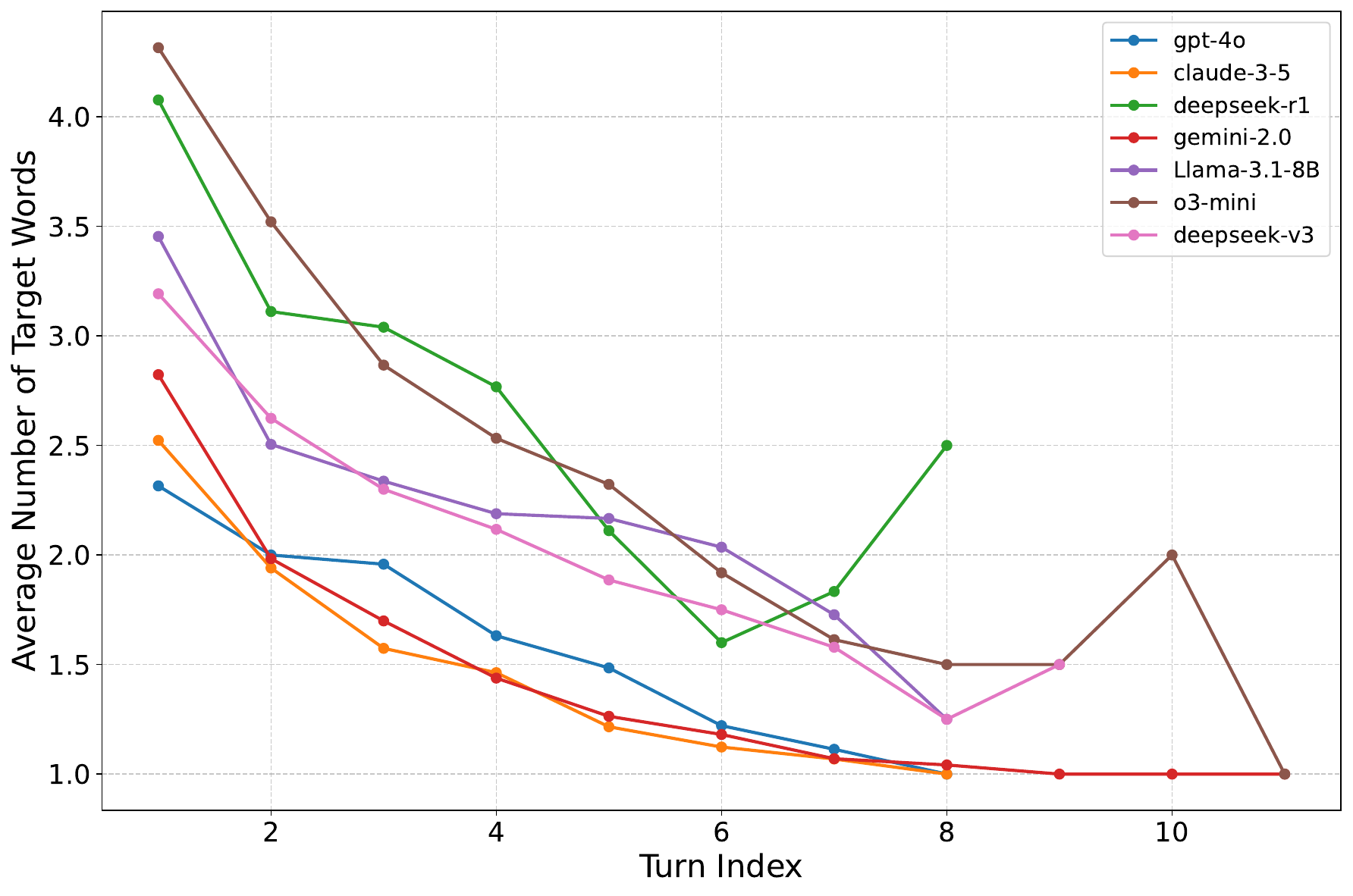}
    \end{minipage}
    \hfill
    \begin{minipage}{0.45\textwidth}
        \centering
        \includegraphics[width=1.0\linewidth]{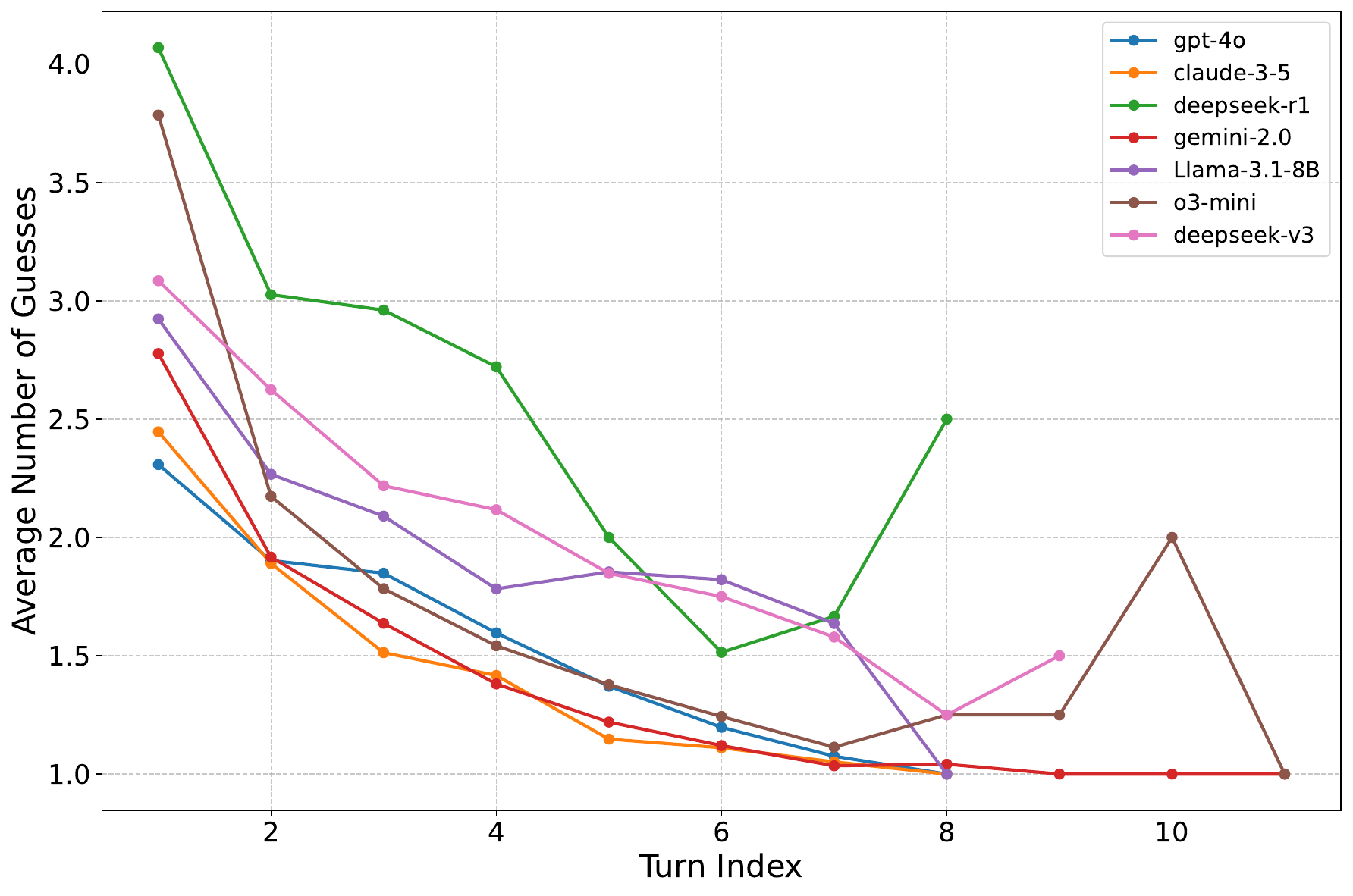}
    \end{minipage}
    \caption{Average number of words targeted (top) and words guessed (bottom) by models at each turn}
        \label{fig:avg_turn}
\end{figure}

\textbf{Number of Targets, Guesses \& Revealed}: in Figure~\ref{fig:avg_turn}, we present the average number of words targeted and words guessed by selected models. We can see that high-performing models such as \textit{o3-mini} and \textit{Deepseek-r1} generate at least 1-2 more words as targets and guesses in the beginning. Targeting and guessing more words in a single turn is the standard strategy in Codenames to win~\cite{spendlove-2022-competitive-language-games-as-creative-tasks}, especially needed when playing against the mock opponent, which reveals one word at each turn. 
Models tend to guess fewer words than were targeted. For instance, \textit{o3-mini} on average targets more than four words but guessed considerably fewer for the first turn, unlike \textit{Deepseek-r1}, which targets and guesses an almost equal number of words. In Figure~\ref{fig:avg_target_guess}, we included the average number of target, guessed and revealed (where the guess is team word) words per model. We can see that only \textit{Deepseek-r1} exceeds the threshold of more than two words (2.2), while the rest have close values (1.5-1.9). It indicates that all models guess wrong words by revealing words from the opponent team or distractors, or even assassin words.

\textbf{Success, Lose \& Aborted Lose Rates}: 
\begin{figure}[ht!]
    \centering
    \begin{minipage}{0.5\textwidth}
        \centering
        \includegraphics[width=0.90\linewidth]{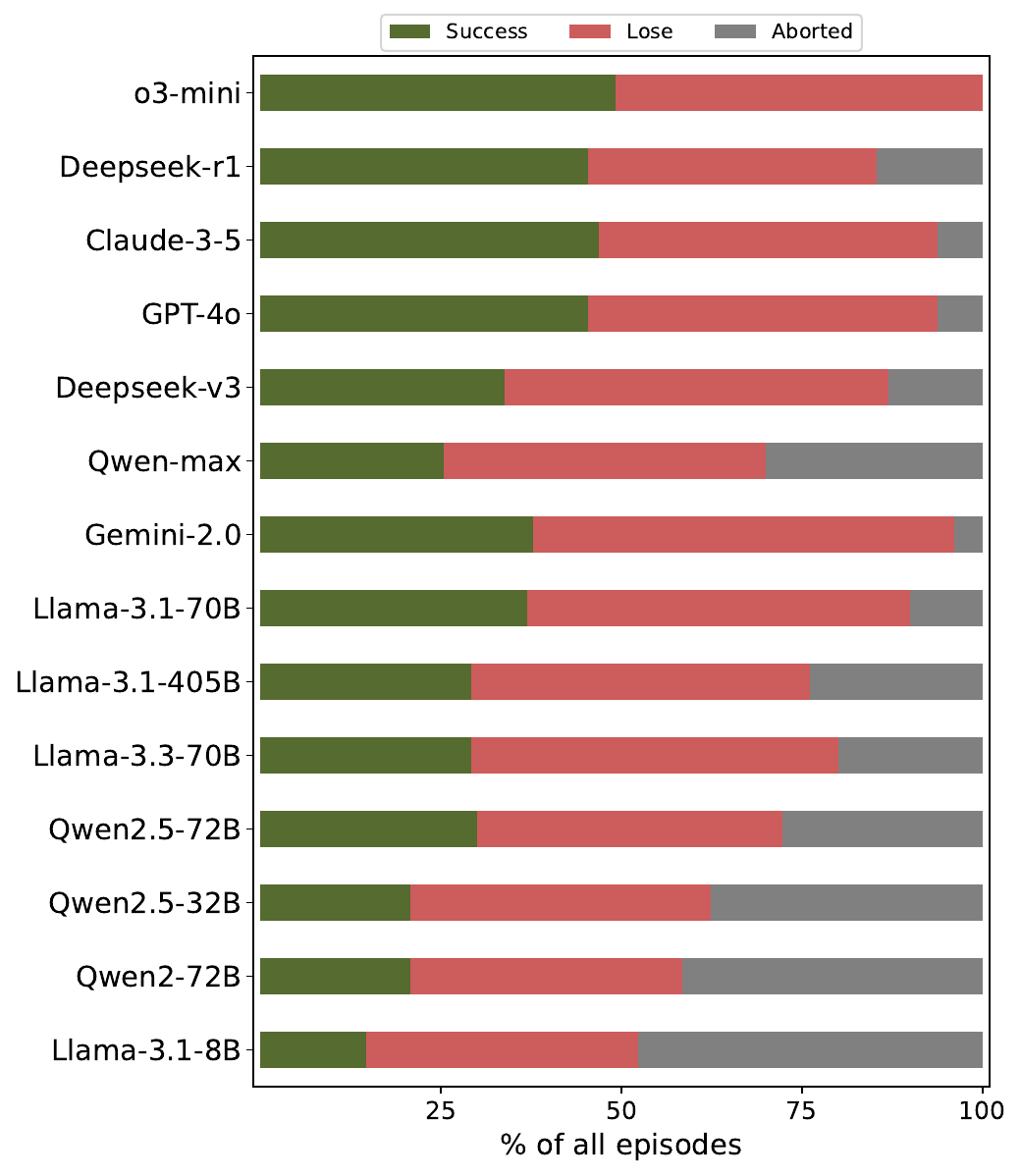}
    \end{minipage}
    
    \hfill
    \begin{minipage}{0.45\textwidth}
        \centering
        \includegraphics[width=1.0\linewidth]{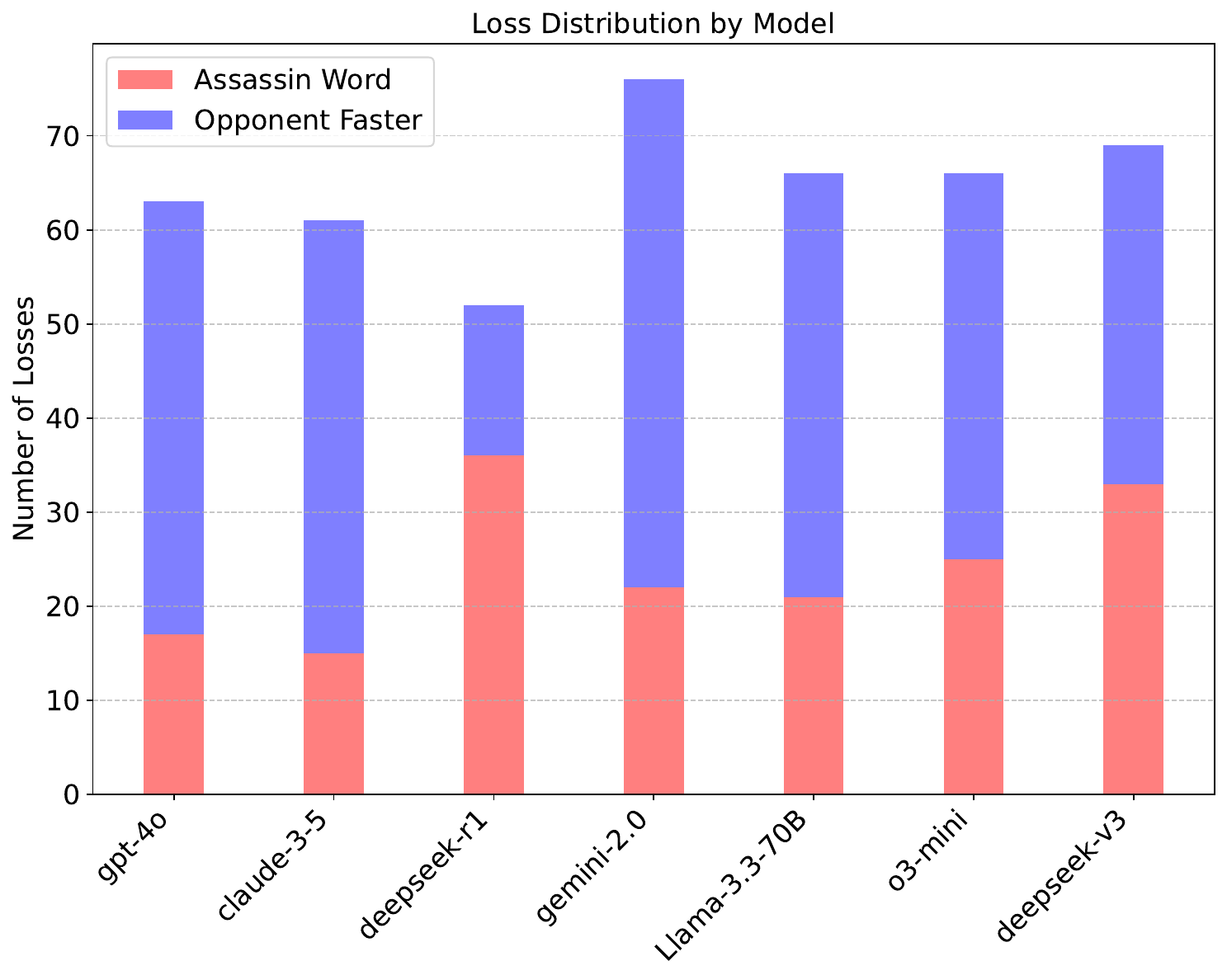}
    \end{minipage}
    \caption{Distribution of Success, Lose, Aborted episodes (up), and distribution of cases where models lose (bottom).}
        \label{fig:success_lose_rates}
\end{figure}
Figure~\ref{fig:success_lose_rates} includes the distribution of episodes across \textit{Success}, \textit{Lose}, and \textit{Aborted}. To recall, \textit{Success} is when a model follows the game's rules and beats the mock opponent by revealing the team words faster, \textit{Lose} is when the mock opponent is faster or when a model reveals assassin words. \textit{Aborted} is when a model does not follow formatting instructions. The top graphic shows that even best-performing models barely reach the 50\% \textit{Success} rates where most episodes are lost or aborted. The ratio of \textit{Aborted} episodes is higher for open-weight models. The bottom graphic divides the \textit{Lost} cases further into two groups: \textit{assassin word is revealed} or \textit{mock opponent is faster}. For most models, the main issue is losing due to being slower in revealing words than the mock opponent. Only \textit{Deepseek-r1} lost more due to revealing more assassin words than others. It shows that all models struggled with the task and lost against a strategy of revealing one word every turn. 

\textbf{Efficiency \& Sensitivity}:
\begin{figure}
    \centering
    \includegraphics[width=1\linewidth]{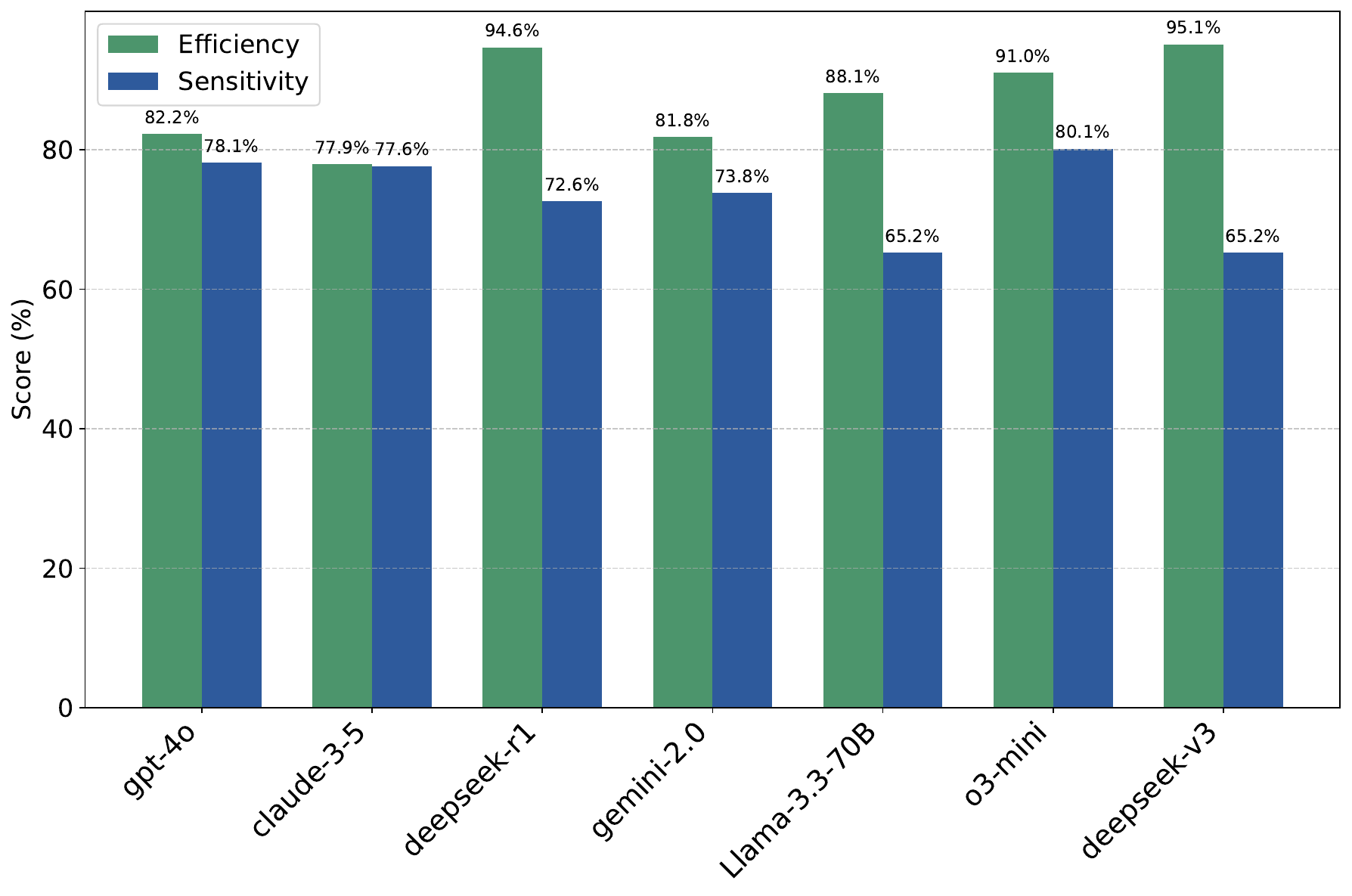}
    \caption{Average \textit{efficiency} and \textit{sensitivity} scores for selected models.}
    \label{fig:efficiency-sensitivity}
\end{figure}
Next, we analyse how efficient the models are regarding targeting multiple words at each turn (see metrics defined in Section~\ref{subsec:metrics}). Figure~\ref{fig:efficiency-sensitivity} shows the \textit{efficiency} and \textit{sensitivity} scores for the selected models. We can observe that \textit{o3-mini}, \textit{Deepseek-r1}, \textit{Llama-3.3-70B} and \textit{Deepseek-v3} have higher efficiency scores, which indicates that these models target two or more words each turn. A similar observation has also been made in Figure~\ref{fig:avg_turn}. By looking at the \textit{sensitivity} scores, we can conclude that \textit{Deepseek-r1} is better at this task than \textit{Deepseek-v3} because it revealed more words (sensitivity score). Models such as \textit{Claude-3.5} and \textit{GPT-4o} are more consistent (efficiency and sensitivity are closer to each other) in terms of the number of targets, guessed, and revealed words.

\begin{table}[ht!]
\centering\footnotesize
\begin{tabular}{|l|c|c|c|c|c|c|}
\hline
\textbf{Model} & \makecell{\textbf{Target} \\ \textbf{Halluc.}} & \makecell{\textbf{Guess} \\ \textbf{Halluc.}} & \makecell{\textbf{Wrong \#} \\\ \textbf{of Guesses}} & \makecell{\textbf{Guess is} \\ \textbf{Clue}} \\ \hline
o3-mini & 0 & 0 & 0 & 0  \\ \hline
DS-r1 & 0 & 0 & 1 & 0  \\ \hline
GPT-4o & 2 & 3 & 0 & 0  \\ \hline
GM-2.0 & 1 & 0 & 4 & 0 \\ \hline
Cl-3.5 & 3 & 5 & 0 & 0  \\ \hline
LM1-70 & 2 & 2 & 1 & 7  \\ \hline
DS-v3 & 6 & 6 & 1 & 2  \\ \hline
LM3-70 & 2 & 2 & 3 & 13 \\ \hline
LM-405 & 10 & 2 & 16 & 0  \\ \hline
QW-72 & 5 & 6 & 0 & 21  \\ \hline
QW-M & 12 & 8 & 0 & 15  \\ \hline
QW-32 & 10 & 7 & 0 & 19  \\ \hline
QW-72B & 9 & 12 & 0 & 30 \\ \hline
LM1-8B & 3 & 7 & 18 & 28  \\ \hline
\end{tabular}
\caption{Error types and their counts for each model where an episode was aborted by the GameMaster.}
\label{table:error_counts}
\end{table}

\textbf{Typical Errors}: To understand where models fail and how higher-performing models differ from lower ones, we analysed the most common errors, then categorised them and counted each occurrence, see Table~\ref{table:error_counts}. 

\begin{figure*}[ht]
    \centering
    \begin{minipage}{0.88\textwidth}
        \centering
        \includegraphics[width=1.0\linewidth]{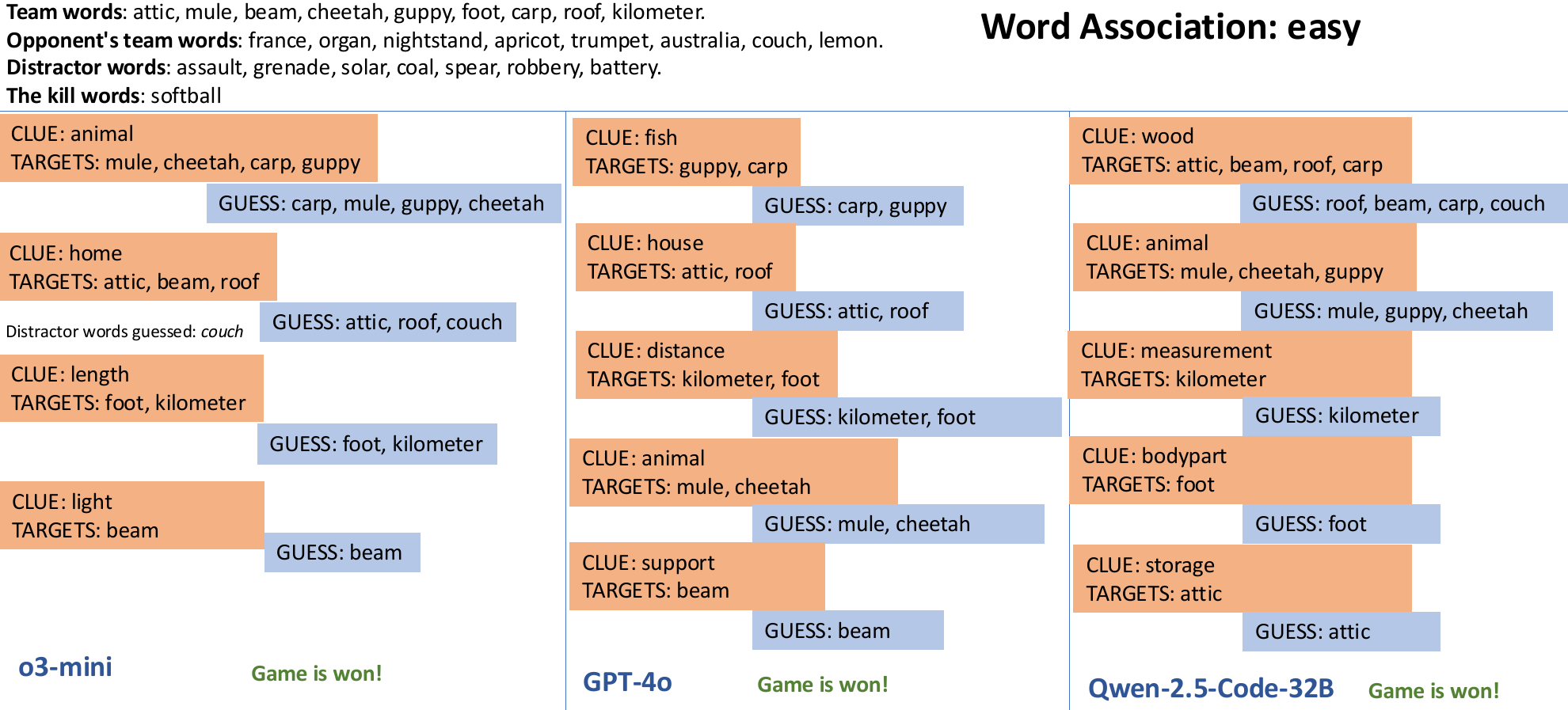}
        \caption{Transcript of an episode in Word Association ``easy'' experiment.}
        \label{fig:qual1}
    \end{minipage}
    \hfill
    \begin{minipage}{0.88\textwidth}
        \centering
        \includegraphics[width=1.0\linewidth]{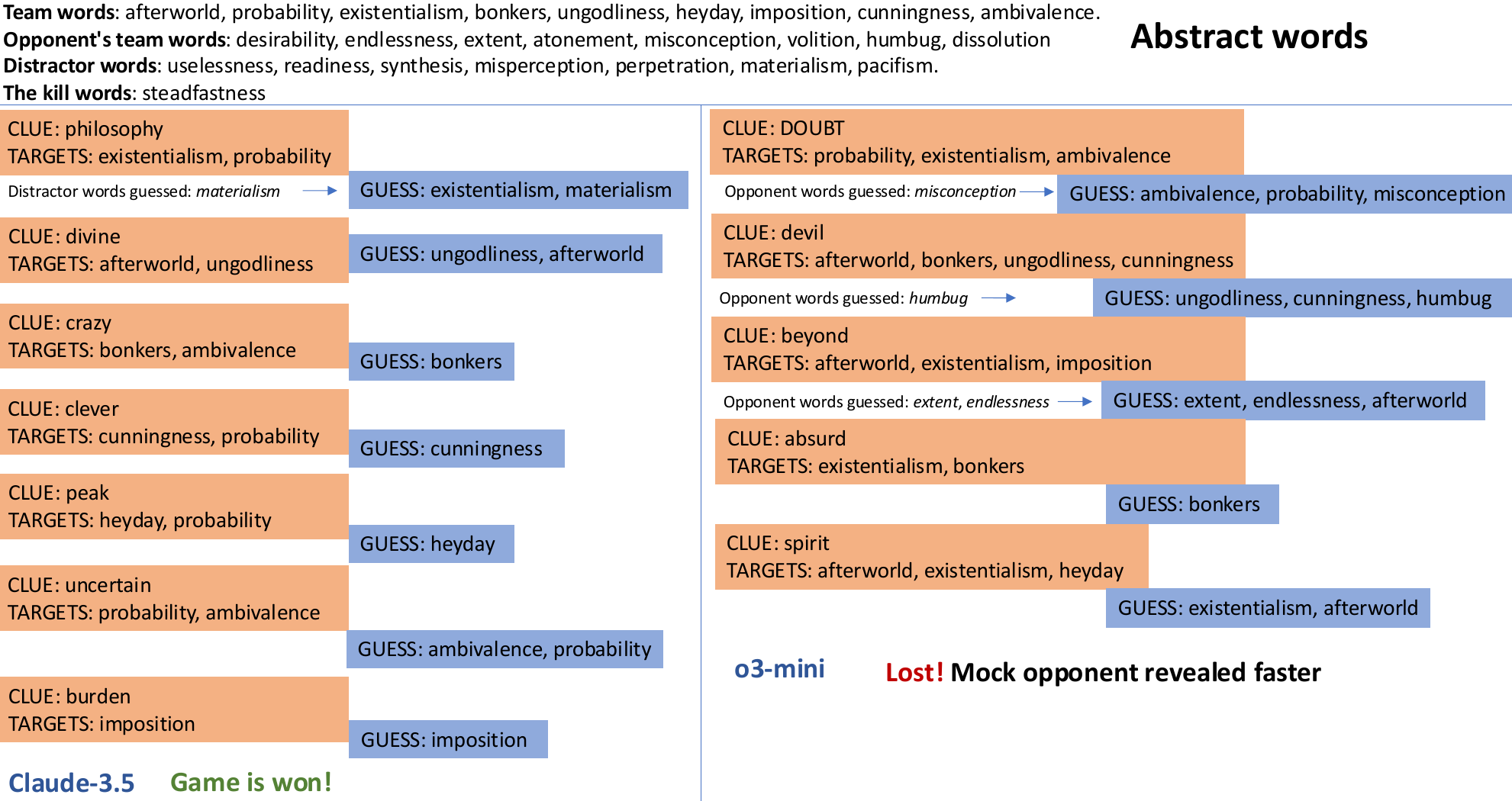}
    \caption{Transcript of an episode in Concreteness ``abstract'' experiment. Note that only the clue is given to player B; the list of targeted words is only to get an insight into the reasoning of player A.}
        \label{fig:qual2}
    \end{minipage}
    
\end{figure*}

The differentiating factor in high-performing models is that hallucination and instruction following issues appear more rarely. For instance, the first error type, \textit{Target Hallucinated}, refers to cases where Spymaster generates a clue and targets some words, but some of these do not exist on the board (as should be known to the model). In such cases, the GameMaster aborts that episode. Similarly, \textit{Guess Hallucinated} is an error that occurs on the Field Operative side where it guesses a word that does not exist on the board. Mostly, \textit{Llama-3.1-405B} and \textit{Llama-3.1-8B} have another issue with guessing the correct number of words that the Spymaster indicates. They tend to guess more than the number of target words (note here: models can guess less but not more than target words). Lastly, the common issue, \textit{Guess is Clue}, with low-performing models is that the guessed word is the same as the clue in many cases. It shows a lack of pragmatic reasoning for choosing unrevealed candidate words from available ones on a board. In all of these cases and some minor ones, e.g.\ tags such as ``CLUE:'', ``TARGETS:'', ``GUESS:'' are omitted, the GameMaster aborts the game because the rules are not followed. Such instruction-following issues happen mostly with \textit{Deepseek-r1} (see Table~\ref{table:error_counts_full}).

\subsection{Qualitative Analysis}

We included sample outputs for the \textit{Word Association - easy} experiment in Figure~\ref{fig:qual1}. Recall that all models achieved the perfect score for this experiment (see Table~\ref{tab:experiment-results}). The words were selected from these categories ``fish'', ``unit of distance'', ``four footed animal'', ``part of a building'', ``fruit'', ``an article of furniture'', ``country'', ``musical instrument'', ``type of fuel'', ``weapon'', ``crime'', ``sport''. 

\textit{o3-mini} generates clues close to the ground truth categories of words. In the second turn, it makes a slight mistake by guessing the distractor word ``couch''. Similarly, \textit{GPT-4o} generates similar clues but focuses on only two words at a time. An interesting case occurs with \textit{Qwen2.5-Coder-32B} where, in the first turn, it targets four words with the clue ``wood''. The other two models targeted the word ``carp'' by choosing the ``fish'' or ``animal'' categories, but \textit{Qwen2.5-Coder-32B} chose the sense of ``carpenter, lumber quality'' to connect the clue ``wood'' to ``carp''.

Figure~\ref{fig:qual2} shows sample outputs for the \textit{Concreteness - abstract} experiment. As we can see, the chosen words are not typical daily life words that would challenge human players in Codenames. \textit{Claude-3.5} manages to play this episode and win the game. We can see that it generates decent clues that combine the target words. It made one mistake by guessing a distractor word in the second turn. The gameplay by \textit{o3-mini} is even more fascinating. The average number of target words is three, and it generates matching clues. However, due to the strategy of targeting and guessing more words, it gives a massive advantage to the opponent by revealing 50\% of their teams' words (``misconception'', ``humbug'', ``extent'', ``endlessness''). Even though the model manages to reveal seven out of nine words (``heyday'' and ``imposition'' were never revealed), it lost the game because the mock opponent revealed words (primarily due to four additional words revealed mistakenly by \textit{o3-mini}).

Figure~\ref{fig:qual3} includes sample episodes for the \textit{Risk level - high} experiment with five assassin words. \textit{o3-mini}, \textit{Claude-3.5}, \textit{Gemini-2.0}, \textit{Deepseek-r1} guessed one of the assassin words and lost the game. \textit{Llama3.3-70B} lost the game due to guessing (six words) more than what was targeted (five words). 

Figure~\ref{fig:qual4} shows samples for the \textit{ambiguous} words. \textit{Deepseek-v3} revealed three opponent words but still managed to win the game.

\section{Discussion}

\textbf{Commercial vs. open}: We can notice that commercial models outperform open-weight ones by some margin. We categorised the errors by models and counted them (see Table~\ref{table:error_counts}). The main reasons for open-weight models having a lower ratio of Played episodes are i) these models often hallucinate while choosing target words, which means they add a word in the target list that does not exist on the board, ii) hallucination also occurs by guessing words that do not exist on the board, iii) guessing the clue word itself. For instance, the performance difference between \textit{Llama3.1-70B} and \textit{405B} can be explained with the bigger model: i) hallucinating target words and ii) guessing too many words.

\textbf{Choice of words}: The selection of words (ambiguous, abstract, high or low frequency, more assassin words) impacts the performance as expected. Of all the experiments, playing against a mock opponent that revealed two words and word associations with difficulty levels proved to be the most challenging. Similarly, abstract words seemed to be more demanding than concrete words. However, we observed that the frequency of words does not directly impact performance when looking at all model results, whereas, for humans, less frequent words might be more challenging. Similar remarks can be made for ambiguity and abstract word sets where the results are somewhat mixed and where humans are expected to find them demanding.

\textbf{Reasoning models}: By looking at the best performing models, we can conclude that the best of one of the commercial and open-weight options are reasoning models where \textit{Deepseek-r1} outperforming some commercial models such as \textit{Gemini-2.0} or \textit{Qwen-max}. However, such an impressive performance comes at the cost of high latency. It took almost two minutes per query for \textit{r1} and two seconds for \textit{v3} (see Table~\ref{tab:latency}).

\textbf{Do LLMs have the required abilities to play Codenames?} The models cannot play efficiently in some experiments by looking at the win rates (Quality Score) for all models. Codenames is a challenging task that involves deep language understanding, theory of mind, cooperation, and pragmatic reasoning. Our experimental results suggest that LLMs do possess knowledge about word associations, and it was shown that they can access it strategically (see Figure~\ref{fig:qual1} where \textit{o3-mini} targets four words with clue ``animal''). Another strategy that we observed is the \textit{risk taking strategy} where models target more than two words per turn to win the game (see Figure~\ref{fig:avg_target_guess}). Such a strategy would be a clear winner against a mock opponent that reveals only one word per turn. However, we have seen cases where this strategy resulted in actually losing the game by revealing the opponent teams' words (see \textit{o3-mini} in Figure~\ref{fig:qual2}). Another risky strategy was observed with the high-risk set, where models could not navigate the experiment with five assassin words. Some models still went on to target a lot of words while risking the error on the guesser side (see \textit{o3-mini} on Figure~\ref{fig:qual4} where it targets nine words at once and loses the game).

The experiments also reveal certain aspects of \textit{pragmatic reasoning} in multi-turn tasks where if a particular clue was not utilised to guess certain target words, it has been revised (see Figure~\ref{fig:qual2} where \textit{o3-mini} targets the word ``existentialism'' with the clue ``doubt'' and it was not guessed, then reintroduced another clue ``spirit'' to the guess the same word again). The cooperation aspect can be seen where some models are consistent in terms of choosing the number of target words and how many of them were correctly guessed (see Figure~\ref{fig:efficiency-sensitivity}, \textit{GPT-4o}, \textit{Claude-3.5}).

\section{Conclusion}

We implemented Codenames to benchmark LLMs by targeting their pragmatic reasoning, language understanding specifically for ad-hoc concept generation, and cooperation capabilities. We tested the most recent commercial and open-weight models on various experiments and difficulty levels. We can generally confirm that commercial models are ahead in performance compared to open-weight ones. The main reasons for better performance can be attributed to having less errors with regards to hallucinations, instruction following, and pragmatic reasoning. However, when looking at played episodes, we can say that even the best performing models do not win over 50\% of the games. It clearly indicates that the task is far from being solved. Overall, the presented solution provides a clear method for benchmarking LLMs using game-based evaluation to target specific capabilities.

\section*{Limitations}

The current study is restricted to only English in its current state. While we have yet to do this, translating the prompts and finding the matching word lists should be possible for other languages, too. We plan to do this in future work.

As discussed in the analysis above, some of the findings are limited to general strategies applied internally by the models. We plan to study the reasoning capabilities in detail to understand the underlying blocks that leads to certain clues or guesses to be generated.

\section*{Ethics Statement}

Using paid proprietary APIs with underlying models about which little is known (training data, model architecture) in academic research is less than ideal. At the moment, the models benchmarked here seem to be the high-performing ones that are commercially used. It is our hope that more open models with high performance will be released soon, and proper research can be done with them.

\bibliography{anthology_0, anthology_1, custom}

\input{appendix}

\end{document}

%% file: appendix.tex
\section{Additional Results}\label{appendix:results}

\begin{table}[h]
    \centering
    \begin{tabular}{l c l}
        
        \textbf{Model} & \textbf{Latency (sec)} & \textbf{Backend} \\ \hline
        
        Llama-3.1-8B & 0.55 & Local \\
        Qwen2.5-32B & 0.67 & Local \\
        GPT-4o & 0.81 & OpenAI \\
        Qwen2-72B & 1.51 & Local \\
        Llama-3.1-70B & 1.48 & Local \\
        Claude-3.5 & 1.28 & Local \\
        Llama-3.1-405B & 1.24 & OpenRouter \\
        Llama-3.3-70B & 1.61 & Local \\
        Qwen2.5-72B & 1.73 & Local \\
        Deepseek-v3 & 2.00 & OpenRouter \\
        Qwen-Max & 4.82 & OpenRouter \\
        o3-mini & 10.91 & OpenAI \\
        Gemini-2.0 & 10.98 & Google \\
        Deepseek-r1 & 111.44 & OpenRouter \\ \hline
        
    \end{tabular}
    \caption{Latencies for benchmarked models.}
    \label{tab:latency}
\end{table}

\begin{table*}[]
\centering
\footnotesize
\begin{tabular}{|l|c|c|c|c|c|c|c|c|c|}
\hline
Model & \makecell{\textbf{Target} \\ \textbf{Hallucinated}} & \makecell{\textbf{Guess} \\ \textbf{Hallucinated}} & \makecell{\textbf{Rambling} \\ \textbf{Error}} & \makecell{\textbf{Repeated} \\ \textbf{Clue}} & \makecell{\textbf{Prefix} \\ \textbf{Error}} & \makecell{\textbf{Wrong \#} \\\ \textbf{of Guesses}} & \makecell{\textbf{Guess} \\ \textbf{is Clue}} &  \makecell{\textbf{Repeated} \\ \textbf{Target}} &  \makecell{\textbf{Double} \\ \textbf{Guess}} \\ \hline
o3-mini & 0 & 0 & 0 & 0 & 0 & 0 & 0 & 0 & 0 \\ \hline
Gemini-2.0 & 1 & 0 & 0 & 0 & 0 & 4 & 0 & 0 & 0 \\ \hline
GPT-4o & 2 & 3 & 0 & 0 & 2 & 0 & 0 & 0 & 0 \\ \hline
Claude-3-5 & 3 & 5 & 0 & 0 & 0 & 0 & 0 & 0 & 0 \\ \hline
Llama-3.1-70B & 2 & 2 & 0 & 0 & 0 & 1 & 7 & 0 & 0 \\ \hline
Deepseek-v3 & 6 & 6 & 0 & 0 & 0 & 1 & 2 & 0 & 0 \\ \hline
Deepseek-r1 & 0 & 0 & 3 & 0 & 14 & 1 & 0 & 0 & 0 \\ \hline
Llama-3.3-70B & 2 & 2 & 3 & 0 & 0 & 3 & 13 & 2 & 0 \\ \hline
Llama-3.1-405B & 10 & 2 & 2 & 0 & 1 & 16 & 0 & 0 & 0 \\ \hline
Qwen2.5-72B & 5 & 6 & 1 & 0 & 0 & 0 & 21 & 1 & 0 \\ \hline
Qwen2.5-7B & 5 & 10 & 6 & 0 & 0 & 0 & 13 & 1 & 1 \\ \hline
Qwen-max & 12 & 8 & 0 & 2 & 0 & 0 & 15 & 2 & 0 \\ \hline
Qwen2.5-32B & 10 & 7 & 1 & 0 & 0 & 0 & 19 & 5 & 1 \\ \hline
Qwen2-72B & 9 & 12 & 0 & 0 & 0 & 0 & 30 & 2 & 0 \\ \hline
Llama-3.1-8B & 3 & 7 & 0 & 0 & 0 & 18 & 28 & 1 & 0 \\ \hline
\end{tabular}
\caption{Error counts for each model}
\label{table:error_counts_full}
\end{table*}

\begin{figure*}[]
    \centering
    \includegraphics[width=1.0\linewidth]{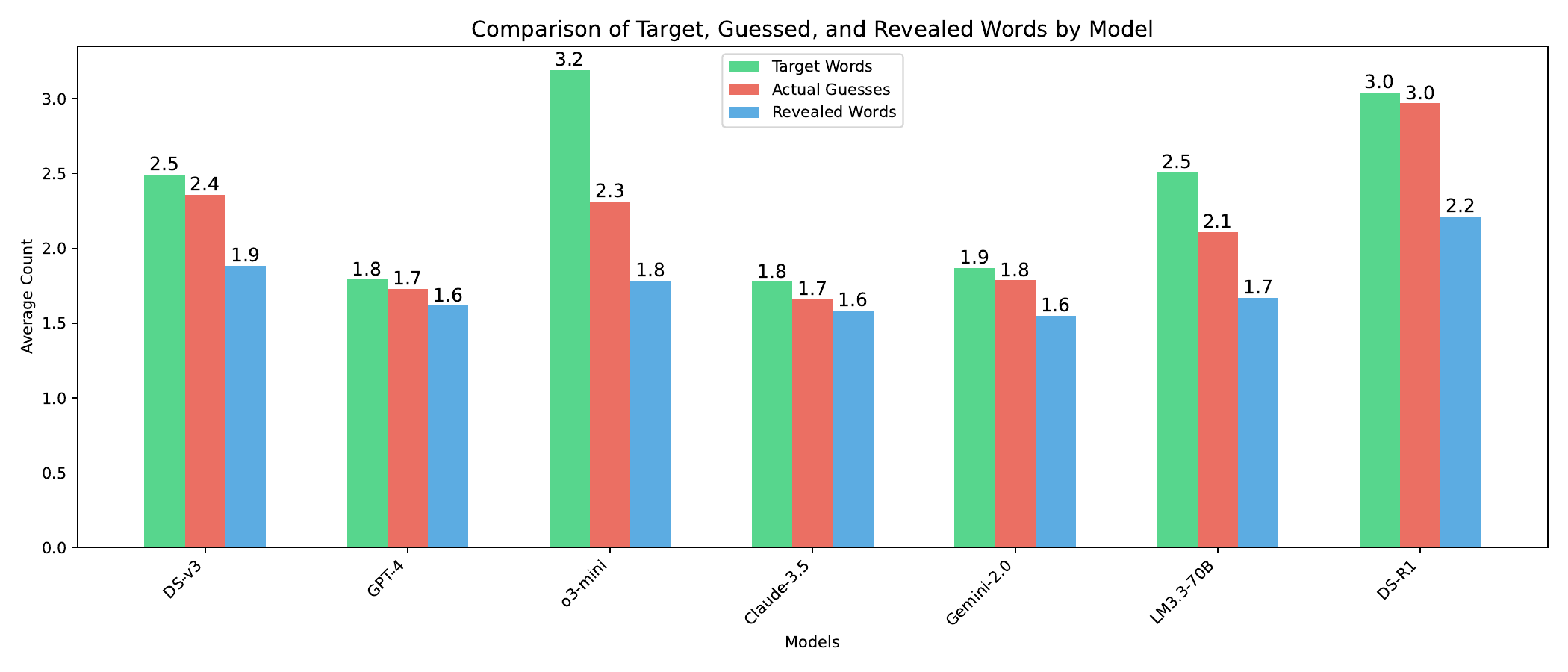}
    \caption{Average number of words that are targeted, guessed, and revealed for selected models.}
    \label{fig:avg_target_guess}
\end{figure*}

\begin{figure*}
    \centering
    \includegraphics[width=1.0\linewidth]{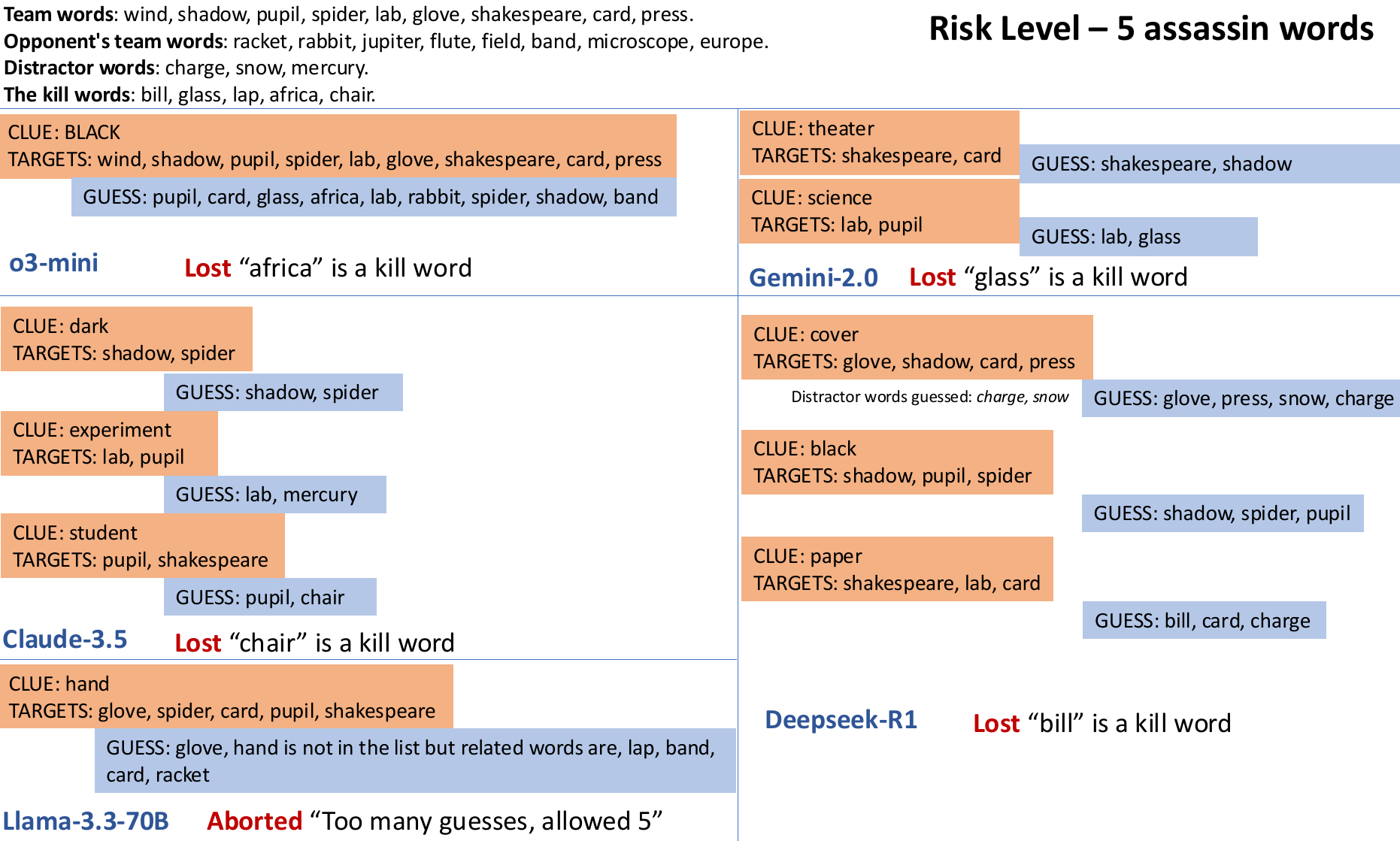}
    \caption{Transcript of an episode in Risk Level ``high'' experiment}
    \label{fig:qual3}
\end{figure*}

\begin{figure*}
    \centering
    \includegraphics[width=1.0\linewidth]{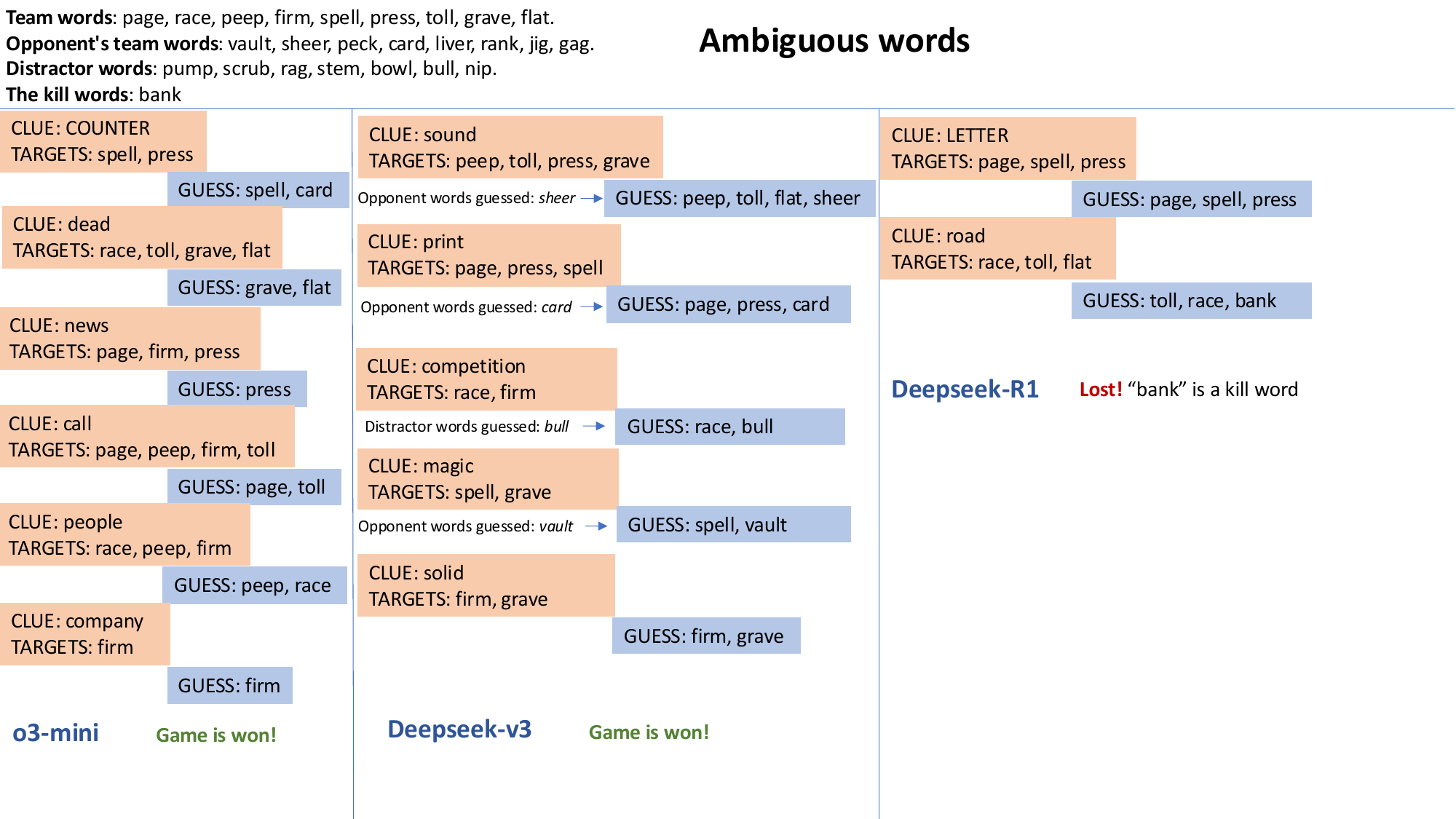}
    \caption{Transcript of an episode in Ambiguity ``ambiguous'' experiment}
    \label{fig:qual4}
\end{figure*}

\clearpage

\section{Prompts}\label{appendix:prompts}

\begin{figure}[ht]
    \centering
    \small
\begin{lstlisting}
Provide one single-word clue that relates to some of your team's words from the list below. You can choose to how many and to which words your clue relates to. The goal is to have your partner guess all of your team's words based on the clues you give them, before the other team has uncovered all of theirs. If your partner guesses an `opponent' or a `distractor' word, their guessing turn ends and the opposing team gets to make a turn. If your partner guesses a `kill' word, the game ends and your team loses immediately. The more words your clue relates to, the quicker you can win the game, but the harder the associations may be to guess for your partner, so choose your clue word wisely. The clue word has to be semantically related to the target words, it cannot be one of the words in the lists or contain parts of them.

Always give your single-word clue and your comma-separated list of related target words in the following format and make your answers as short as possible, never include any other text than is required in this form:

CLUE: <WORD>
TARGETS: <TARGETS>

Your team words are: $team_words.
Your opponent's team words are: $opponent_words.
Distractor words are: $innocent_words.
The kill words are: $assassin_words.
\end{lstlisting}
\caption{Spymaster Prompt}
\label{fig:spymaster-prompt}
\end{figure}

\begin{figure}[ht]
    \centering
    \small
\begin{lstlisting}
Provide a comma-separated list of up to $number words from the following list that best relate or are most closely associated with the word `$clue'. Always start your list of guess(es) with `GUESS: ' and do not include any other text in your answer.

$board
\end{lstlisting}
\caption{Field Operative Prompt}
\label{fig:field-operative-prompt}
\end{figure}

\begin{figure}[ht]
    \centering
    \footnotesize
\begin{lstlisting}
The words $correct_guesses were guessed correctly. The word $correct_guess was guessed correctly. The word $incorrect_guess was guessed but is an $assignment word. Your teammate's turn ended there.
\end{lstlisting}
\caption{Spymaster Feedback}
\label{fig:spymaster-feedback}
\end{figure}

\begin{figure}[ht]
    \centering
    \footnotesize
\begin{lstlisting}
The words $correct_guesses were guessed correctly. The word $correct_guess was guessed correctly. The word $incorrect_guess was guessed but is an $assignment word. 

Your turn ended there.
\end{lstlisting}
\caption{Field Operative Feedback}
\label{fig:field-operative-feedback}
\end{figure}

\begin{figure}[ht]
    \centering
    \footnotesize
\begin{lstlisting}
Now provide another clue relating to some of your remaining team words and a list of the related target words. Remember to start your clue with `CLUE: ', put a new line, and start your comma-separated list of target words with `TARGETS: '. Notice: some words have been removed from the lists compared to previous requests.

Your remaining team words are: $team_words.
Remaining words for your opponent are: $opponent_words.
Remaining distractor words are: $innocent_words.
Remaining kill words are: $assassin_words.
\end{lstlisting}
\caption{Intermittent Spymaster Prompt}
\label{fig:intermittent-spymaster}
\end{figure}

\begin{figure}[ht]
    \centering
    \footnotesize
\begin{lstlisting}
Now provide another comma-separated list of at least 1 and up to $number words from the following list of words that best relate or are most closely associated with the word `$clue'. Remember to start your answer with `GUESS: '. Notice: some words have been removed from the list compared to previous requests.

$board
\end{lstlisting}
\caption{Intermittent Field Operative Prompt}
\label{fig:intermittent-field-operative}
\end{figure}

%% file: arxiv_version.bbl
\begin{thebibliography}{33}
\providecommand{\natexlab}[1]{#1}

\bibitem[{Beekhuizen et~al.(2021)Beekhuizen, Armstrong, and
  Stevenson}]{beekhuizen2021ambiguity}
Barend Beekhuizen, Blair~C. Armstrong, and Suzanne Stevenson. 2021.
\newblock \href {https://doi.org/10.1111/COGS.12943} {Probing lexical
  ambiguity: Word vectors encode number and relatedness of senses}.
\newblock \emph{Cogn. Sci.}, 45(5).

\bibitem[{Beyer et~al.(2024)Beyer, Chalamalasetti, Hakimov, Madureira, Sadler,
  and Schlangen}]{beyer2024clembench2024challengingdynamiccomplementary}
Anne Beyer, Kranti Chalamalasetti, Sherzod Hakimov, Brielen Madureira, Philipp
  Sadler, and David Schlangen. 2024.
\newblock \href {https://arxiv.org/abs/2405.20859} {clembench-2024: A
  challenging, dynamic, complementary, multilingual benchmark and underlying
  flexible framework for llms as multi-action agents}.
\newblock \emph{Preprint}, arXiv:2405.20859.

\bibitem[{Bitton et~al.(2022)Bitton, Guetta, Yosef, Elovici, Bansal, Stanovsky,
  and Schwartz}]{bitton-2022-winogavil}
Yonatan Bitton, Nitzan~Bitton Guetta, Ron Yosef, Yuval Elovici, Mohit Bansal,
  Gabriel Stanovsky, and Roy Schwartz. 2022.
\newblock \href
  {http://papers.nips.cc/paper\_files/paper/2022/hash/a96fe863f85c59789bba63588a9557b4-Abstract-Datasets\_and\_Benchmarks.html}
  {Winogavil: Gamified association benchmark to challenge vision-and-language
  models}.
\newblock In \emph{Advances in Neural Information Processing Systems 35: Annual
  Conference on Neural Information Processing Systems 2022, NeurIPS 2022, New
  Orleans, LA, USA, November 28 - December 9, 2022}.

\bibitem[{Brysbaert and New(2009)}]{subtlexus}
Marc Brysbaert and Boris New. 2009.
\newblock \href
  {https://www.ugent.be/pp/experimentele-psychologie/en/research/documents/subtlexus}
  {{SUBTLEXus}: American word frequencies based on subtitle corpora}.
\newblock Accessed: 2025-02-12.

\bibitem[{Brysbaert et~al.(2014)Brysbaert, Warriner, and
  Kuperman}]{brysbaert2014concreteness}
Marc Brysbaert, Amy~Beth Warriner, and Victor Kuperman. 2014.
\newblock Concreteness ratings for 40 thousand generally known english word
  lemmas.
\newblock \emph{Behavior research methods}, 46:904--911.

\bibitem[{Castro et~al.(2021)Castro, Curley, and
  Hertzog}]{castro2021category-norms}
Nichol Castro, Taylor Curley, and Christopher Hertzog. 2021.
\newblock Category norms with a cross-sectional sample of adults in the united
  states: Consideration of cohort, age, and historical effects on semantic
  categories.
\newblock \emph{Behavior Research Methods}, 53:898--917.

\bibitem[{Chalamalasetti et~al.(2023)Chalamalasetti, G{\"o}tze, Hakimov,
  Madureira, Sadler, and Schlangen}]{chalamalasetti-etal-2023-clembench}
Kranti Chalamalasetti, Jana G{\"o}tze, Sherzod Hakimov, Brielen Madureira,
  Philipp Sadler, and David Schlangen. 2023.
\newblock \href {https://doi.org/10.18653/v1/2023.emnlp-main.689} {clembench:
  Using game play to evaluate chat-optimized language models as conversational
  agents}.
\newblock In \emph{Proceedings of the 2023 Conference on Empirical Methods in
  Natural Language Processing}, pages 11174--11219, Singapore. Association for
  Computational Linguistics.

\bibitem[{Chv{\'a}til and
  Ku{\v{c}}erovsk{\`y}(2015)}]{chvatil-2015-codenames-game}
Vlaada Chv{\'a}til and Tom{\'a}{\v{s}} Ku{\v{c}}erovsk{\`y}. 2015.
\newblock \emph{Codenames}.
\newblock Czech Games Edition.

\bibitem[{Cserh{\'a}ti et~al.(2022)Cserh{\'a}ti, Kollath, Kicsi, and
  Berend}]{cserhati-2022-codenames-co-occurrence-counting}
R{\'e}ka Cserh{\'a}ti, Istvan Kollath, Andr{\'a}s Kicsi, and G{\'a}bor Berend.
  2022.
\newblock \href {https://doi.org/10.18653/v1/2022.cmcl-1.5} {Codenames as a
  game of co-occurrence counting}.
\newblock In \emph{Proceedings of the Workshop on Cognitive Modeling and
  Computational Linguistics}, pages 43--53, Dublin, Ireland. Association for
  Computational Linguistics.

\bibitem[{de~Rijk and
  Marecek(2020)}]{deRijk-2020-paper-word-embeddings-collocations}
Micha de~Rijk and David Marecek. 2020.
\newblock \href {http://ufal.mff.cuni.cz/pbml/114/art-de-rijk-marecek.pdf}
  {Using word embeddings and collocations for modelling word associations}.
\newblock \emph{Prague Bull. Math. Linguistics}, 114:35.

\bibitem[{de~Rijk(2020)}]{deRijk-2020-thesis-codenames-modelling-word-association}
Micha Theo~Neri de~Rijk. 2020.
\newblock Codenames: a practical application for modelling word association.
\newblock Master's thesis, Charles University, Prague, Czech Republic.

\bibitem[{DeepSeek-AI et~al.(2025)DeepSeek-AI, Guo, Yang, Zhang, Song, and
  et~al.}]{deepseekai2025deepseekr1incentivizingreasoningcapability}
DeepSeek-AI, Daya Guo, Dejian Yang, Haowei Zhang, Junxiao Song, and Ruoyu~Zhang
  et~al. 2025.
\newblock \href {https://arxiv.org/abs/2501.12948} {Deepseek-r1: Incentivizing
  reasoning capability in llms via reinforcement learning}.
\newblock \emph{Preprint}, arXiv:2501.12948.

\bibitem[{DeepSeek-AI et~al.(2024)DeepSeek-AI, Liu, Feng, Xue, Wang, Wu, Lu,
  and et~al.}]{deepseekv3}
DeepSeek-AI, Aixin Liu, Bei Feng, Bing Xue, Bingxuan Wang, Bochao Wu, Chengda
  Lu, and et~al. 2024.
\newblock \href {https://arxiv.org/abs/2412.19437} {Deepseek-v3 technical
  report}.
\newblock \emph{Preprint}, arXiv:2412.19437.

\bibitem[{Deng et~al.(2024)Deng, Zhao, Tang, Gerstein, and
  Cohan}]{deng-etal-2024-investigating}
Chunyuan Deng, Yilun Zhao, Xiangru Tang, Mark Gerstein, and Arman Cohan. 2024.
\newblock \href {https://doi.org/10.18653/v1/2024.naacl-long.482}
  {Investigating data contamination in modern benchmarks for large language
  models}.
\newblock In \emph{Proceedings of the 2024 Conference of the North American
  Chapter of the Association for Computational Linguistics: Human Language
  Technologies (Volume 1: Long Papers)}, pages 8706--8719, Mexico City, Mexico.
  Association for Computational Linguistics.

\bibitem[{Golchin and Surdeanu(2024)}]{DBLP:conf/iclr/GolchinS24}
Shahriar Golchin and Mihai Surdeanu. 2024.
\newblock \href {https://openreview.net/forum?id=2Rwq6c3tvr} {Time travel in
  llms: Tracing data contamination in large language models}.
\newblock In \emph{The Twelfth International Conference on Learning
  Representations, {ICLR} 2024, Vienna, Austria, May 7-11, 2024}.
  OpenReview.net.

\bibitem[{Grattafiori et~al.(2024)Grattafiori, Dubey, Jauhri, Pandey, Kadian,
  and et~al.}]{llama31}
Aaron Grattafiori, Abhimanyu Dubey, Abhinav Jauhri, Abhinav Pandey, Abhishek
  Kadian, and et~al. 2024.
\newblock \href {https://arxiv.org/abs/2407.21783} {The llama 3 herd of
  models}.
\newblock \emph{Preprint}, arXiv:2407.21783.

\bibitem[{Jaramillo et~al.(2020)Jaramillo, Charity, Canaan, and
  Togelius}]{jaramillo-2020-transformers-for-word-association-in-codenames}
Catalina~M. Jaramillo, Megan Charity, Rodrigo Canaan, and Julian Togelius.
  2020.
\newblock \href {https://ojs.aaai.org/index.php/AIIDE/article/view/7435} {Word
  autobots: Using transformers for word association in the game codenames}.
\newblock In \emph{Proceedings of the Sixteenth {AAAI} Conference on Artificial
  Intelligence and Interactive Digital Entertainment, {AIIDE} 2020, virtual,
  October 19-23, 2020}, pages 231--237. {AAAI} Press.

\bibitem[{Kim et~al.(2019)Kim, Ruzmaykin, Truong, and
  Summerville}]{kim-2019-understanding-NLP-via-codenames}
Andrew Kim, Maxim Ruzmaykin, Aaron Truong, and Adam Summerville. 2019.
\newblock \href {https://ojs.aaai.org/index.php/AIIDE/article/view/5239}
  {Cooperation and codenames: Understanding natural language processing via
  codenames}.
\newblock In \emph{Proceedings of the Fifteenth {AAAI} Conference on Artificial
  Intelligence and Interactive Digital Entertainment, {AIIDE} 2019, October
  8-12, 2019, Atlanta, Georgia, {USA}}, pages 160--166. {AAAI} Press.

\bibitem[{Koyyalagunta et~al.(2021)Koyyalagunta, Sun, Draelos, and
  Rudin}]{koyyalagunta-2021-codenames-language-graphs}
Divya Koyyalagunta, Anna~Y. Sun, Rachel~Lea Draelos, and Cynthia Rudin. 2021.
\newblock \href {https://doi.org/10.1613/JAIR.1.12665} {Playing codenames with
  language graphs and word embeddings}.
\newblock \emph{J. Artif. Intell. Res.}, 71:319--346.

\bibitem[{Kumar et~al.(2021)Kumar, Steyvers, and
  Balota}]{kumar-2021-connector-asociative-vs-distributional-semantic-models}
Abhilasha~Ashok Kumar, Mark Steyvers, and David~A. Balota. 2021.
\newblock \href {https://doi.org/10.1111/COGS.13053} {Semantic memory search
  and retrieval in a novel cooperative word game: {A} comparison of associative
  and distributional semantic models}.
\newblock \emph{Cogn. Sci.}, 45(10).

\bibitem[{Lu et~al.(2023)Lu, Bigoulaeva, Sachdeva, Madabushi, and
  Gurevych}]{lu-2023-emergence-just-in-context-learning}
Sheng Lu, Irina Bigoulaeva, Rachneet Sachdeva, Harish~Tayyar Madabushi, and
  Iryna Gurevych. 2023.
\newblock \href {https://doi.org/10.48550/ARXIV.2309.01809} {Are emergent
  abilities in large language models just in-context learning?}
\newblock \emph{CoRR}, abs/2309.01809.

\bibitem[{Ozturkler et~al.(2023)Ozturkler, Malkin, Wang, and
  Jojic}]{ozturkler-2023-thinksum}
Batu Ozturkler, Nikolay Malkin, Zhen Wang, and Nebojsa Jojic. 2023.
\newblock \href {https://doi.org/10.18653/V1/2023.ACL-LONG.68} {Thinksum:
  Probabilistic reasoning over sets using large language models}.
\newblock In \emph{Proceedings of the 61st Annual Meeting of the Association
  for Computational Linguistics (Volume 1: Long Papers), {ACL} 2023, Toronto,
  Canada, July 9-14, 2023}, pages 1216--1239. Association for Computational
  Linguistics.

\bibitem[{Qwen et~al.(2025)Qwen, Yang, Yang, Zhang, Hui, and et~al.}]{qwen25}
Qwen, An~Yang, Baosong Yang, Beichen Zhang, Binyuan Hui, and et~al. 2025.
\newblock \href {https://arxiv.org/abs/2412.15115} {Qwen2.5 technical report}.
\newblock \emph{Preprint}, arXiv:2412.15115.

\bibitem[{Shaikh et~al.(2023)Shaikh, Ziems, Held, Pariani, Morstatter, and
  Yang}]{shaikh-2023-cross-cultural-pragmatic-inference-with-codenames-duet}
Omar Shaikh, Caleb Ziems, William Held, Aryan~J. Pariani, Fred Morstatter, and
  Diyi Yang. 2023.
\newblock \href {https://doi.org/10.18653/V1/2023.FINDINGS-ACL.410} {Modeling
  cross-cultural pragmatic inference with codenames duet}.
\newblock In \emph{Findings of the Association for Computational Linguistics:
  {ACL} 2023, Toronto, Canada, July 9-14, 2023}, pages 6550--6569. Association
  for Computational Linguistics.

\bibitem[{Song et~al.(2024)Song, Wang, Li, and
  Lin}]{song2024goodbadgreedyevaluation}
Yifan Song, Guoyin Wang, Sujian Li, and Bill~Yuchen Lin. 2024.
\newblock \href {https://arxiv.org/abs/2407.10457} {The good, the bad, and the
  greedy: Evaluation of llms should not ignore non-determinism}.
\newblock \emph{Preprint}, arXiv:2407.10457.

\bibitem[{Spendlove and
  Ventura(2022)}]{spendlove-2022-competitive-language-games-as-creative-tasks}
Brad Spendlove and Dan Ventura. 2022.
\newblock \href
  {http://computationalcreativity.net/iccc22/papers/CCC-2022\_paper\_73.pdf}
  {Competitive language games as creative tasks with well-defined goals}.
\newblock In \emph{Proceedings of the 13th International Conference on
  Computational Creativity, Bozen-Bolzano, Italy, June 27 - July 1, 2022},
  pages 291--299. Association for Computational Creativity {(ACC)}.

\bibitem[{Spendlove and Ventura(2023)}]{spendlove-constraints-for-creativity}
Brad Spendlove and Dan Ventura. 2023.
\newblock \href
  {https://computationalcreativity.net/iccc23/papers/ICCC-2023_paper_144.pdf}
  {Constraints as catalysts: A (de) construction of codenames as a creative
  task}.
\newblock In \emph{Proceedings of the 14th International Conference on
  Computational Creativity}.

\bibitem[{Srivastava et~al.(2022)Srivastava, Rastogi, Rao, Shoeb, Abid, Fisch,
  Brown, Santoro, Gupta, Garriga{-}Alonso, and
  et~al.}]{srivastava-2022-bigbench}
Aarohi Srivastava, Abhinav Rastogi, Abhishek Rao, Abu Awal~Md Shoeb, Abubakar
  Abid, Adam Fisch, Adam~R. Brown, Adam Santoro, Aditya Gupta, Adri{\`{a}}
  Garriga{-}Alonso, and et~al. 2022.
\newblock \href {https://doi.org/10.48550/ARXIV.2206.04615} {Beyond the
  imitation game: Quantifying and extrapolating the capabilities of language
  models}.
\newblock \emph{CoRR}, abs/2206.04615.

\bibitem[{Stephenson et~al.(2024)Stephenson, Sidji, and
  Ronval}]{DBLP:journals/corr/abs-2412-11373}
Matthew Stephenson, Matthew Sidji, and Beno{\^{\i}}t Ronval. 2024.
\newblock \href {https://doi.org/10.48550/ARXIV.2412.11373} {Codenames as a
  benchmark for large language models}.
\newblock \emph{CoRR}, abs/2412.11373.

\bibitem[{Wei et~al.(2022)Wei, Tay, Bommasani, Raffel, Zoph, Borgeaud,
  Yogatama, Bosma, Zhou, Metzler, Chi, Hashimoto, Vinyals, Liang, Dean, and
  Fedus}]{wei-2022-emergence}
Jason Wei, Yi~Tay, Rishi Bommasani, Colin Raffel, Barret Zoph, Sebastian
  Borgeaud, Dani Yogatama, Maarten Bosma, Denny Zhou, Donald Metzler, Ed~H.
  Chi, Tatsunori Hashimoto, Oriol Vinyals, Percy Liang, Jeff Dean, and William
  Fedus. 2022.
\newblock \href {https://openreview.net/forum?id=yzkSU5zdwD} {Emergent
  abilities of large language models}.
\newblock \emph{Trans. Mach. Learn. Res.}, 2022.

\bibitem[{Wu et~al.(2024)Wu, Tang, Mitchell, and Li}]{DBLP:conf/iclr/WuTML24}
Yue Wu, Xuan Tang, Tom~M. Mitchell, and Yuanzhi Li. 2024.
\newblock \href {https://openreview.net/forum?id=S2oTVrlcp3} {Smartplay : {A}
  benchmark for llms as intelligent agents}.
\newblock In \emph{The Twelfth International Conference on Learning
  Representations, {ICLR} 2024, Vienna, Austria, May 7-11, 2024}.
  OpenReview.net.

\bibitem[{Yang et~al.(2024)Yang, Yang, Hui, Zheng, and et~al.}]{qwen2}
An~Yang, Baosong Yang, Binyuan Hui, Bo~Zheng, and et~al. 2024.
\newblock \href {https://arxiv.org/abs/2407.10671} {Qwen2 technical report}.
\newblock \emph{Preprint}, arXiv:2407.10671.

\bibitem[{Zhou et~al.(2024)Zhou, Zhu, Mathur, Zhang, Yu, Qi, Morency, Bisk,
  Fried, Neubig, and Sap}]{DBLP:conf/iclr/Zhou0MZYQMBFNS24}
Xuhui Zhou, Hao Zhu, Leena Mathur, Ruohong Zhang, Haofei Yu, Zhengyang Qi,
  Louis{-}Philippe Morency, Yonatan Bisk, Daniel Fried, Graham Neubig, and
  Maarten Sap. 2024.
\newblock \href {https://openreview.net/forum?id=mM7VurbA4r} {{SOTOPIA:}
  interactive evaluation for social intelligence in language agents}.
\newblock In \emph{The Twelfth International Conference on Learning
  Representations, {ICLR} 2024, Vienna, Austria, May 7-11, 2024}.
  OpenReview.net.

\end{thebibliography}
